\pdfoutput=1

\documentclass[11pt]{article}

\usepackage[]{acl}

\usepackage{times}
\usepackage{latexsym}

\usepackage{bm}
\usepackage{color}
\usepackage{amssymb}
\usepackage{amsmath}
\usepackage{amssymb}
\usepackage{amsmath}
\usepackage{algorithm}  
\usepackage[noend]{algorithmic}
\usepackage[switch]{lineno}
 \usepackage{xcolor}
 \usepackage{color}
 \usepackage{bbding}
 \usepackage{graphicx}
 \usepackage{booktabs}
 \usepackage{caption}
 \usepackage{subcaption}
 \usepackage{multirow}
\usepackage[T1]{fontenc}

\usepackage[utf8]{inputenc}

\usepackage{microtype}
\usepackage{hyperref}

\title{A Hierarchical Interactive Network for Joint Span-based \\ 
Aspect-Sentiment Analysis}

\author{Wei Chen$^{1,2}$, Jinglong Du$^{3}$, Zhao Zhang$^{4}$, Fuzhen Zhuang$^{2,5\ast}$, Zhongshi He$^{1}$\!\thanks{~~Corresponding authors.} \\
         $^{1}$College of Computer Science, Chongqing University, Chongqing, China \\
         $^{2}$Institute of Artificial Intelligence, Beihang University, Beijing 100191, China \\
         $^{3}$College of Medical Informatics, Chongqing Medical University, Chongqing, China \\
         $^{4}$Institute of Computing Technology, Chinese Academy of Sciences, Beijing, China \\
         $^{5}$SKLSDE, School of Computer Science, Beihang University, Beijing 100191, China\\
     \{mlg\_cwei,zshe\}@cqu.edu.cn, zhuangfuzhen@buaa.edu.cn}

\begin{document}
\maketitle
\begin{abstract}
Recently, some span-based methods have achieved encouraging performances for joint aspect-sentiment analysis, which first extract aspects (aspect extraction) by detecting aspect boundaries and then classify the span-level sentiments (sentiment classification). However, most existing approaches either sequentially extract task-specific features, leading to insufficient feature interactions, 
or they encode aspect features and sentiment features in a parallel manner, implying that feature representation in each task is largely independent of each other except for input sharing. Both of them ignore the internal correlations between the aspect extraction
 and sentiment classification. 
To solve this problem,
we novelly propose a hierarchical interactive network (HI-ASA) to 
model two-way interactions between two tasks appropriately, where the hierarchical interactions involve two steps: shallow-level interaction and deep-level interaction. 
First, we utilize cross-stitch mechanism to combine the different task-specific features selectively as the input to ensure proper two-way interactions. Second, the mutual information technique is applied to mutually constrain learning between two tasks in the output layer, thus the aspect input and the sentiment input are capable of encoding features of the other task via backpropagation. Extensive experiments on three real-world datasets demonstrate HI-ASA's superiority over baselines\footnote{The source codes are released in \url{https://github.com/cwei01/HI-ASA}.}. 
\end{abstract}

\section{Introduction}
\label{intro}
Aspect-sentiment analysis (ASA)~\cite{yan2021unified,birjali2021comprehensive} aims at extracting all the aspects and their corresponding sentiments within the text simultaneously.
And it can be divided into two tasks, i.e., aspect extraction (AE) and sentiment classification (SC). AE is to extract the
aspects~\cite{jakob2010extracting,poria2016aspect,karimi2021adversarial} in the sentence, and SC aims to predict the
sentiments~\cite{jiang2011target,lin2019deep,karimi2021adversarial} 
for the extracted aspects.

In recent years, the span-based models  \cite{hu2019open,zhou2019span,lin2020shared,lv2021span} are increasingly becoming an alternative for ASA because of their inherent advantages (e.g., they can avoid sentiment inconsistency and huge search space problems in tagging-based models~\cite{luo2019doer,wang2021end}), where the aspects are extracted by directly predicting the boundary distributions, and the sentiment polarities are classified based on the aspect-level words.
For example, in the sentence ``The screen size is satisfactory but the phone battery capacity is limited.", the aspect spans are ``screen size'' and ``the phone battery'', and span-level sentiments are positive and negative, respectively.
Formally, most of the existing methods
can be divided into two types: sequential
encoding~\cite{jebbara2016aspect,zhou2019span} and parallel encoding~\cite{hu2019open,lv2021span} due to different ways in encoding task-specific
features.  
In sequential encoding, the task-specific features are extracted sequentially, 
i.e., features extracted later have no direct associations with previous ones, which is a unidirectional interaction strategy.
In parallel encoding, task-specific features
are extracted independently except for using shared input, i.e., the interaction is
only present in input sharing. Hence, both encoding methods above fail to model two-way interactions between AE and SC appropriately.

In practice, the learning of the AE and SC may mutually influence each other.
On the one hand, sentiment words can be understood better if given the desired aspects. For example, in the two sentences: ``The battery has a \textbf{large} capacity.'' and ``The computer case is too \textbf{large} to carry.'', we can find that the sentiment word ``large'' expresses positive sentiment when describing the aspect ``hard drive'' while negative sentiment towards ``computer''.
This actually shows that incorporating the aspect features is conducive to SC.
On the other hand, since sentimental expressions are usually close to aspects, the potential sentiment features may provide useful signals for AE.
For example, if we observe the word ``\textbf{spicy}'' appears in a restaurant review, it is likely to exist an aspect related to food. Thus, incorporating the sentiment feature also facilitates aspect detection.  These two examples illustrate that it is essential to reasonably establish the two-way associations between two tasks.

Inspired by the above analysis, in this work, we propose a novel Hierarchical Interactive model for joint span-based Aspect-Sentiment Analysis (HI-ASA).
Specifically, the hierarchical interactions are achieved in two steps: shallow-level interaction and deep-level interaction. The former learns semantic-level interactive features to facilitate information sharing between two tasks in encoding layer. And the latter takes advantage of the technique of mutual information maximization to ensure two-way interaction between the two tasks.
Extensive experimental comparisons against state-of-the-art
solutions demonstrate the effectiveness of our proposed methods.

\begin{figure*}[t]
	\centering
	\setlength{\fboxrule}{0.pt}
	\setlength{\fboxsep}{0.pt}
	\fbox{
		\includegraphics[width=0.7\linewidth]{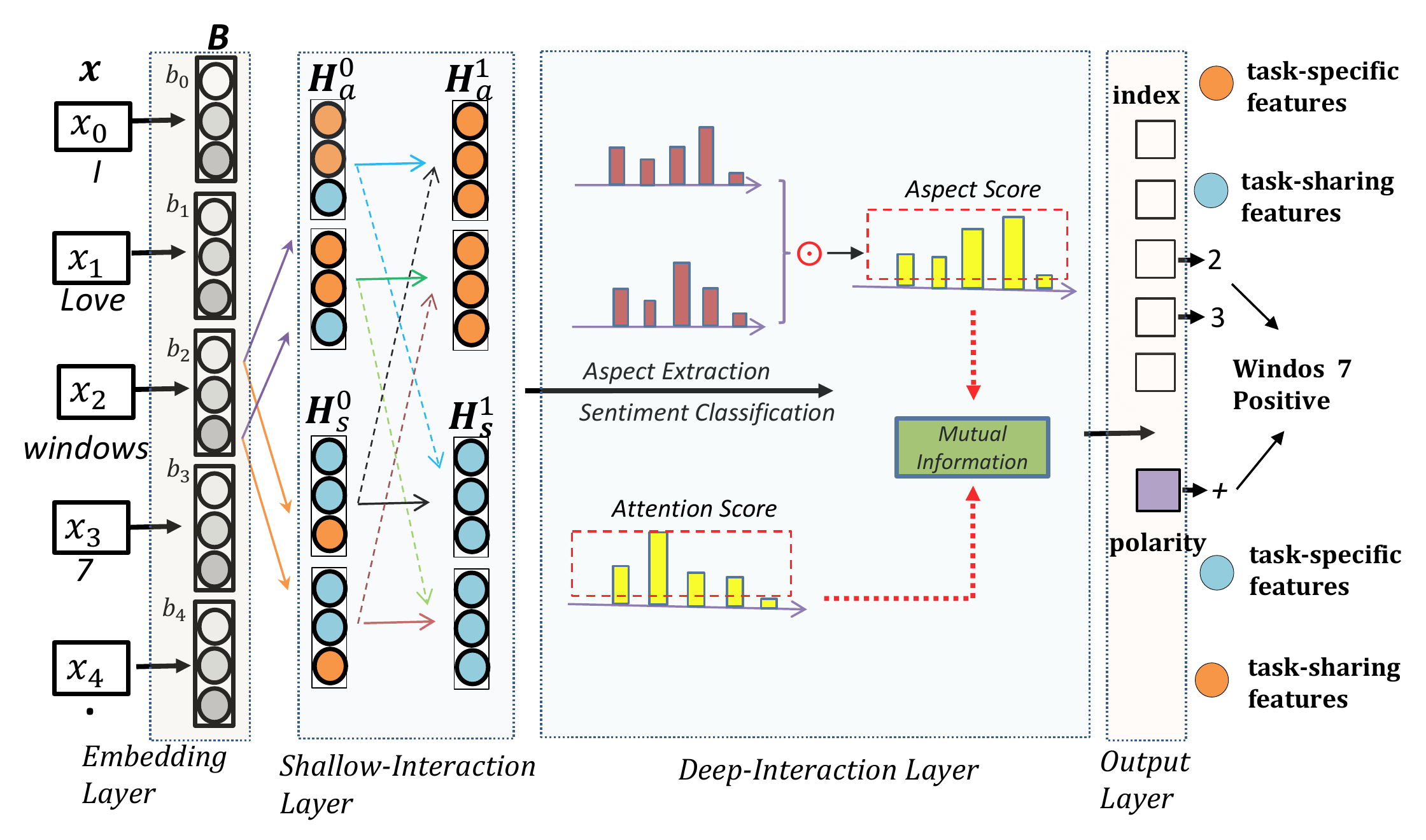}}
	\caption{The architecture of our HI-ASA.
	}
	\label{model}
\end{figure*}
\section{Model}

  In this section, we formally define the problem of joint span-based aspect-sentiment analysis.
  Giving a training set $\{(\bm{x}_i, \{(s_i^j, e_i^j, p_i^j)\}_{j=1}^{l_i})\}_{i=1}^N$, in each sample $(\bm{x}_i, \{(s_i^j, e_i^j, p_i^j)\}_{j=1}^{l_i})$, $\bm{x}_i=\{{x}_{i,1}, {x}_{i,2}, \ldots, {x}_{i,n_i}\}$ is a $n_i$-length sentence, and $(s_i^j, e_i^j, p_i^j)$ identifies an aspect with $s_i^j$ and $e_i^j$ as the start and end boundaries, and $p_i^j$ as sentiment polarity. $l_i$ is the number of aspects in the sentence.

  The overall architecture is illustrated in Figure~\ref{model}, which mainly consists of two parts: AE and SC. 
 We model the hierarchical interactions between two tasks appropriately to enhance the correlations.
For the input section,
  firstly, we leverage BERT~\cite{devlin2018bert} to extract the semantic information for input sentence. The output through the  transformer layers is $\bm{B}=\{b_{1}, b_{2}, \ldots, b_{n}\}\in \mathbb{R}^{n\times d}$, where $d$ is the BERT's embedding size, $n$ is the length of sentence. Then, we stack GRUs upon $\bm{B}$ for different tasks, here, we define the outputs of AE and SC as ${\bm{H}}_{a}^{0}$   $\in$ $\mathbb{R}^{n\times  \hat{d} }$ and ${\bm{H}}_{s}^{0}$  $\in$ $\mathbb{R}^{n\times  \hat{d} }$, respectively, where $\hat{d}$ is the dimension size of GRU.  
  
  \subsection{Shallow-Level Interaction}
  In previous models~\cite{hu2019open,lin2020shared,lv2021span}, the feature representations are extracted independently except for using shared input. 
  In other words, two tasks have no associations with each other, which is not in line with human ituition.
  Intuitively, the features of SC are not only derived from its own features of previous layers, but also come from the features of AE, and vice versa. Therefore, we design a shallow-level interaction strategy inspired by the idea of cross-stitch mechanism~\cite{misra2016cross}.
  Its core is to selectively combine the different features to reasonably model two-way interactions. Specifically,
  the calculation is as follows:
  
  {\small
\begin{equation}\left[\begin{array}{c}
        {\bm{H}}^{1}_a \\
        {\bm{H}}^{a}_s
    \end{array}\right]=\left[\begin{array}{cc}
        \bm{\gamma }_{\mathrm{aa}} & \bm{\alpha}_{\mathrm{sa}} \\
        \bm{\alpha}_{as} & \bm{\gamma}_{\mathrm{ss}}
    \end{array}\right]\left[\begin{array}{c}
        {\bm{H}}^{0}_a \\
        {\bm{H}}^{0}_s
    \end{array}\right]
    \label{titch}
\end{equation} }
where $\bm{\gamma}_{aa}$, $\bm{\gamma}_{ss}$ are the task-specific parameters, and $\bm{\alpha}_{sa}$, $\bm{\alpha}_{as}$ are task-sharing parameters. 
${\bm{H}}_{a}^{1}$ and ${\bm{H}}_{s}^{1}$ are the output features of AE and SC in the last encoding layer, respectively.
We can observe that the cross-stitch unit is beneficial for the interaction between AE and SC features.
To speed up the training process, we define the following constraints, (1): $\bm{\gamma}_{aa} = \bm{\gamma}_{ss}$ , $ \bm{\alpha}_{sa} = \bm{\alpha}_{as}$
(2): $\bm{\gamma}_{aa} + \bm{\alpha}_{sa} =1 $.
Thus, the Eq.\ref{titch} is simplified to:

{\small
\begin{equation}
    \begin{aligned}
    \boldsymbol{H}_{a}^{1}=\alpha \cdot \boldsymbol{H}_{s}^{0}+(1-\alpha) \cdot \boldsymbol{H}_{a}^{0} \\
    \boldsymbol{H}_{s}^{0}=\alpha \cdot \boldsymbol{H}_{a}^{0}+(1-\alpha) \cdot \boldsymbol{H}_{s}^{0} 
    \label{cross}
    \end{aligned}
\end{equation}
}
where $\alpha$ controls information  transferred from the other task.

   \subsection{Deep-Level Interaction}
\begin{figure}[t]
  \centering
  \setlength{\fboxrule}{0.pt}
  \setlength{\fboxsep}{0.pt}
  \fbox{
  \includegraphics[width=1\linewidth]{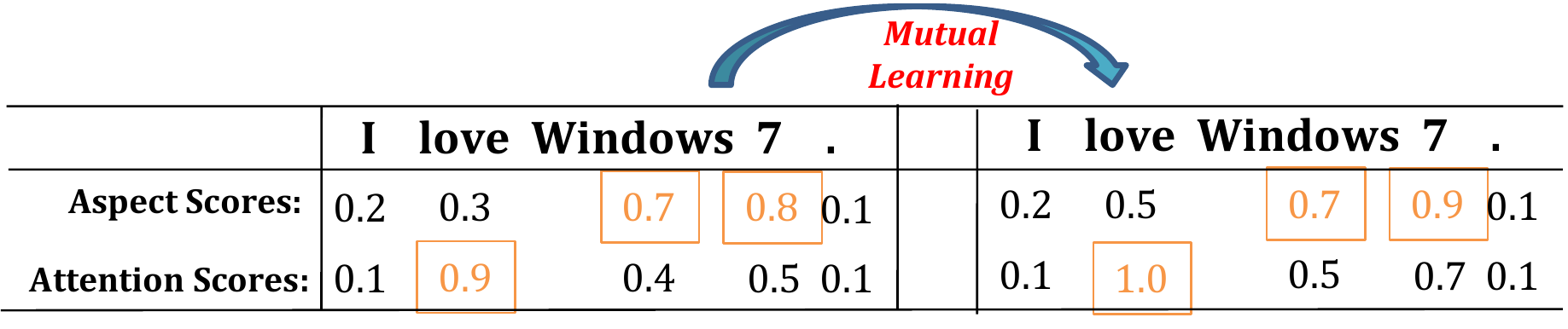}
  }
  \caption{The example of our mutual information maximization strategy in sentence ``I love Windows 7.''.
  }
  \label{example}
  \vspace{-10pt}
  \end{figure}

   In the last section, we model the two-way interactions from the encoding layer. Although some semantic-level associations can be captured, we believe that task-level associations cannot be adequately modeled.
   To better understand our proposed
method, we give an intuitive explanation in Figure~\ref{example}.
   We observe the span-based aspect extracion models put higher scores on the entity words~(e.g., ``Windows'' ,``7'' ). But with aspect-level sentiment learning, higher
attention weights are not only put on some polarity words~(e.g., ``love''), but also on the entity words. Therefore, it would be interesting if we perform aspect-sentiment mutual learning on both tasks.
  Thus, we propose the deep-level interactive strategy based on mutual information maximization~\cite{kong2019mutual}, where the task-level information is shared by the two tasks in the output layer.
 We will explain in detail how to obtain the aspect scores  (i.e., the more a word acts like an aspect word, the higher its score) and attention scores for these two tasks.

  \textbf{Aspect Extraction ($\mathcal{E}$).}
  \label{ae}
 We extract the aspects by predicting the boundaries, i.e., start position and end position. 
 Specifically, we employ a linear classifier to predict the start scores $g_{s} =  \operatorname{sigmoid}(\bm{V_{s}} \bm{H}_a^{1}  )$, i.e., $\bm{V_{s}}$ is a trainable vector, and end scores $g_{e}$ can be derived in a similar manner.
 During training phase, we define the learning object:
 $\mathcal{J}_{ae} \!\!=\!\! -\sum_{i=1}^{n} {\bm{p}}_{s}^i\log{(\bm{g^i}_{s})} -\sum_{i=1}^{n}{\bm{p}}_{e}^i\log{(\bm{g^i}_{e})}$, where $\bm{p}_{s}^i$ and $\bm{p}_{e}^i$ are the ground truths of the boundaries. $\bm{g}_{s}^i$ and $\bm{g}_{e}^i$ are the predicted distributions.
 
  For the aspect score $\mathcal{E}$, we cannot use the boundary distributions of aspect (i.e., $g_s$ and $g_e$) directly due to their different meanings. In fact, the span of aspect is relatively short, we can use the average boundary distributions to approximate the aspect scores, so we have: $\mathcal{E}=\frac{\operatorname{pooling}\left({g_{s}}\right)+ \operatorname{pooling}\left({g_{e}}\right)}{2}$,
  where $\operatorname{pooling(\cdot)}$ is the mean-pooling function, in our work, our set window size as $1 \times 3$. 
  Therefore, suppose the start score $g_{s}$ is [0.2,0.3,0.7,0.9,0.1], the first item of $\operatorname{pooling}\left({g_{s}}\right)$ is $\frac{0+0.2+0.3}{3} = 0.17$, and the $\operatorname{pooling}\left({g_{s}}\right)$ is 
  [0.17,0.4,0.63,0.57,0.33], such strategy can guarantee that the aspect distribution approximates the boundary distributions as much as possible.
  
   \textbf{Sentiment Classification ($\mathcal{A}$).}
   \label{sc}
  In aspect-level sentiment classification, the attention weights of words produced during the training towards the specific aspects could reveal word-polarity information. Thus, we can  leverage the attention distribution as the sentiment features to help detect the boundaries of aspects. 
  To this end, we choose the over-and-over attention model~\cite{huang2018aspect} to implement the sentiment classification. Here, we directly define the attention distribution as $\mathcal{A}$, (i.e., $\mathcal{A}$ is attention scores in Figure~\ref{example}). During training phase, the optimization goal for SC is:
  $\mathcal{J}_{sc}=-\sum_{i=1}^{N} \boldsymbol{y}_{i}^{T} \log \hat{\boldsymbol{y}}_{i}
  $, where $\hat{\bm{y}}_{i}$ is the predicted sentiment distribution, and ${\bm{y}}_{i}^T$ is the corresponding ground truth, $N$ is the number of samples.

\textbf{Mutual Information Maximization.}
  After getting the aspect distribution $\mathcal{E}$ and sentiment attention distribution $\mathcal{A}$. 
  As discussed above, it should be interesting if we perform aspect-sentiment mutual learning on both tasks.
  In Figure~\ref{example}, we can find that the aspect scores and attention weights are both improved after applying the mutual learning technique.
  We need to maximize the similarity between two distributions during the training. An intuitive idea is to use Kullback Leibler (KL) divergence to measure the distance between distributions, considering the correlations are bidirectional, we define the following similarity measurement metrics using Jenson Shannon (JS) divergence:
  
  {\small
  \begin{equation}
    JS(\mathcal{E} \| \mathcal{A})=\frac{1}{2} K L\left(\mathcal{E} \| \frac{\mathcal{E}+\mathcal{A}}{2}\right)+\frac{1}{2} K L\left(\mathcal{A} \| \frac{\mathcal{E}+\mathcal{A}}{2}\right)
\end{equation}}
where the JS divergence is symmetric, thus we can leverage it to constrain two tasks to learn from each other, resulting in balanced interaction
between AE and SC. Whereas in sequential and
parallel encoding, sentiment features have no direct
impact on the information of aspect features.
Finally, considering the mutual information maximization between two tasks, HI-ASA is optimaled by combining three sections: 
\begin{equation}
    \vspace{-0.2cm}
   \underset{\theta}{\arg \min } \left(\mathcal{J}_{ae} +\mathcal{J}_{sc} + \beta \cdot J S(\mathcal{E} \| \mathcal{A}) \right)
  \label{all_loss}
  \vspace{-0.15cm}
\end{equation} 
where $\theta$ is the set of model's parameters, $\mathcal{J}_{ae}$ and $\mathcal{J}_{sc}$ are the optimization goals of AE and SC respectively. $\beta$ is a balanced parameter. During the test phase, HI-ASA outputs two parts, one is the scores of boundaries, based on which we leverage the heuristic algorithm~\cite{lin2020shared} to extract all the aspects; the other one is corresponding sentiment polarities of extracted aspects.
    \begin{table}[t]
  	\renewcommand{\arraystretch}{0.8}
  	\setlength{\tabcolsep}{0.4mm}{
  		\begin{tabular}{cccccc}
  			\hline
  			Dataset & { \#Sentences} & {\#Aspects} & { \#+} & {\#-} & \#0 \\ \hline
  			\textit{Restaurant} & 3900 & 6603 & 4134 & 1538 & 931 \\
  			\textit{Laptop} & 1869 & 2936 & 1326 & 900 & 620 \\
  			\textit{Tweets} & 2350 & 3243 & 703 & 274 & 2266 \\    \hline
  	\end{tabular}}
  	\caption{Statistics of the datasets. ``+/-/0''  refer to the positive, negative, and neutral sentiment classes.}
  	\label{statistic}
  \end{table}

    \begin{table}[t]
        \renewcommand{\arraystretch}{0.6}
        \setlength{\tabcolsep}{0.4mm}{
              \begin{tabular}{c|c|c|c}
        \hline 
                                    Span-based  Models       & \begin{tabular}[c]{@{}c@{}}\textit{Laptop}\\ \end{tabular} & \begin{tabular}[c]{@{}c@{}}\textit{Res}\\ \end{tabular} & \begin{tabular}[c]{@{}c@{}}\textit{Tweets}\\ \end{tabular} \\ \hline
         
          Zhou~\cite{zhou2019span} & 59.76                                                 & 71.98 & 51.44                                               
             \\ Hu-pipeline~\cite{hu2019open}& 68.06                                                & 74.92 & 57.69 \\ Hu-joint~\cite{hu2019open}& 64.59                                                & 72.47 & 54.55 \\
             
                                                                                              Hu-collapsed\cite{hu2019open}& 
                                48.66                  & 57.85                                                   & 
                                48.11     
                                                                        \\
                                                                         SPRM~\cite{lin2020shared}& 
                                68.72                 & 79.17                                                  & 
                                59.45     \\
                                 S-AESC~\cite{lv2021span}& 
                                65.88                 & 74.18                                                  & 
                                54.73    \\
                                 HI-ASA & 
                                \textbf{70.39}                & \textbf{79.90}                                                  & 
                                \textbf{60.36}     \\
                                 ~~~~~~~~ - w/o SI& 
                                68.05                 & 79.10                                                & 
                                60.04     \\
                                                                                          ~~~~~~~~ - w/o DI       & 67.45                                       &    79.50                                         & 60.13                                        \\ \hline 
            \end{tabular}}
            \caption{The performance (F1-score) comparisons with different methods. Baseline results are retrieved from published papers.}
            \label{main}
            \vspace{-5pt}
          \end{table}
\section{Experiments}
\subsection{Datasets and Settings}

 \textbf{Datasets.} In our experiments, we use three public datasets, including \textit{Laptop}~\cite{pontiki2014semeval}, \textit{Restaurant}~\cite{pontiki2014semeval,pontiki2015semeval,pontiki2016semeval} and \textit{Tweets}~\cite{mitchell2013open}, which have been widely used in previous works~\cite{lv2021span}. 
  The statistic details of experimental dataset are refer to Table~\ref{statistic}. For each sentence in datasets, the gold span boundaries and sentiment polarity labels are available. Specifically, \textit{Restaurant} is the union set of the restaurant domain from SemEval2014, SemEval2015 and SemEval2016. \textit{Laptop} contains costumer reviews in the electronic product domain, which is collected from SemEval Challenge. \textit{Tweets} is composed of twitter posts from different users.
  
   \textbf{Settings.}
  In the experiments, the commonly used metric F1-score (F1) is selected to evaluate for aspect-sentiment analysis, accuracy is applied to sentiment classification.
  A correct predicted aspect only when it matches the gold aspect and the corresponding polarity.
  We utilize the BERT-Large model as the backbone network,where the number of transformers is 24 and the hidden size is 784.
  In addition, we use Adam optimizer with a learning rate of 3e-5, the batch size is 32 and the dropout probability of 0.1 is used. 

\subsection{Main Results}

The comparisons between HI-ASA and the baselines are presented in Table \ref{main}. 
Specifically, ``- w/o SI'' means we remove the shallow-interaction layer, ``- w/o DI'' denotes removing the deep-interaction layer.
We can observe: \textbf{(1)} Overall, our proposed HI-ASA consistently achieves the best F1 scores across all the baselines. More specifically, compared to the state-of-the-art approach SPRM, HI-ASA improves the performance by about 1.67\%, 0.63\%, and 0.91\% on three datasets, respectively. These observations
indicate the carefully designed HI-ASA is capable of achieving better performances. The reason can be concluded two folds: one is that we selectively combine the task-specific features of each task to reasonably model two-way interactions in the encoding layer; the second benefit is that we model a balanced task-level interactions. Under this framework, aspect and sentiment associations are captured appropriately in a hierarical manner.  
\textbf{(2)}  Besides, we investigate the effectiveness of each single module, i.e., ``- w/o SI'' and ``- w/o DI''. We can see that when a certain module is removed, the performance of our model decreases, which indicates the indispensability of each module.

\subsection{Ablation Study}
 \begin{table}[t]
	\renewcommand{\arraystretch}{0.6}
	\setlength{\tabcolsep}{0.4mm}{
		\begin{tabular}{c|c|c|c}
			\hline 
			Aspect Extraction            & \begin{tabular}[c]{@{}c@{}}\textit{Laptop}\\ \end{tabular} & \begin{tabular}[c]{@{}c@{}}\textit{Res}\\ \end{tabular} & \begin{tabular}[c]{@{}c@{}}\textit{Tweets}\\ \end{tabular} \\ \hline
			
			Hu \cite{hu2019open}   & 83.35                                                 & 82.38                                                     & 75.28 \\
			
			SPRM \cite{lin2020shared}    & 84.72                                                & 86.71                                                   & 69.85 \\
			S-AESC \cite{lv2021span}    & 85.19                                                & 84.20                                                   & 76.04 \\
			
			HI-ASA & \textbf{86.30}                                                & 86.93                                                     & \textbf{76.36}                                                 \\
			~~~~~~~~ - w/o SI&84.50                        & \textbf{87.33}                                         &                                     75.81 \\
			~~~~~~~~ - w/o DI       & 85.10                                                & 87.09                                           & 75.99                                               \\ \hline 
			Sentiment Classifiction       & \textit{Laptop}                                              & \textit{Res}                                        & \textit{Tweets}                                              \\ \hline 
			Hu \cite{hu2019open}    & 81.39                                                 & 89.95                                                     & 75.16 \\
			SPRM \cite{lin2020shared}    & 81.50                                                 & 90.35                                                     & 78.34 \\
			HI-ASA   & \textbf{85.02}                                                 & \textbf{93.18}                                                     & \textbf{83.50} \\
			~~~~~~~~ - w/o SI   & 84.07                                                & 92.53                                                    & 82.90 \\
			
			~~~~~~~~ - w/o DI & 83.75                                                 & 92.62                                                     & 83.49                                \\ \hline

			\hline
			
	\end{tabular}}
	\caption{The top part is the performance (F1-score) comparisons with different methods on aspect extraction. And the bottom part is the performance (Accuracy) comparisons with different methods on sentiment classifiction.}
	\label{ablation}
\end{table}
\begin{table}[t]
	\renewcommand{\arraystretch}{0.6}
	\setlength{\tabcolsep}{0.6mm}{
		\begin{tabular}{ccccccc}
			\hline
			Dataset$/$$\alpha$ & 0 & 0.1 & 0.2 & 0.3 & 0.4 & 0.5  \\ \hline
			\textit{Laptop}        & 68.05&
			\textbf{70.39}&
			67.62&
			68.63&
			68.58&
			70.36
			\\
			\textit{Restaurant}            & 79.14&
			78.93&
			79.73&
			79.45&
			\textbf{79.90}&
			78.48
			
			\\
			\textit{Tweet}         & 60.04&
			\textbf{60.36}&
			59.70&
			59.89&
			58.79&
			58.19  \\ \hline
	\end{tabular}}
	\caption{The performance (F1-score) of different sharing ratios ($\alpha$) on three datasets.}
	\label{case}
\end{table}
In this section, we go deeper into HI-ASA and analyse the results on both tasks. The
results are shown in Table~\ref{ablation}.
Generally, we observe that HI-ASA outperforms baseline competitors on both tasks, which indicates the effectiveness of the two-way interaction on both aspect extraction and sentiment classification.
For the AE task, HI-ASA can enhance the performance of most of the baselines. 
On the SC task, HI-ASA outperforms SPRM by 3.98\%, 2.83\%, and 5.16\% on three datasets, respectively, which implies that it is more able to boost the performance of SC compared to AE.

\subsection{ Parameter  Analysis $\alpha$}
In HI-ASA, the parameter $\alpha$ controls information transfer between two tasks.
For a special case, if $\alpha$ = 0, our approach will degenerate to the parallel encoding. We tune $\alpha$ in the range of [0.0,0.1,0.2,0.3,0.4,0.5]
and the results are presented in Table~\ref{case}.
It is worth noting that we set the value of $\alpha$ below 0.5. This setting is inspired by the ituition that the features for each task should come more from the task itself, rather than the other task.
We can notice the best results are achieved when $\alpha$ is 0.1, 0.4, and 0.1 on three datasets, respectively, rather than 0.0. This actually shows our interaction is successful. 

Furthermore, we find an interesting phenomenon is that $\alpha$ is small (i.e., 0.1) on \textit{Laptop} and \textit{Tweets}, and large on \textit{Restaurant} when performing the best performance. We conjecture the reason lies in that \textit{Restaurant} has more samples than the other two datasets, requiring more knowledge interactions for a better learning process.
\section{Related Work}
Aspect-sentiment Analysis (ASA)~\cite{yan2021unified,birjali2021comprehensive} is an essential task in sentiment analysis and can be separated into two tasks, i.e., aspect extraction (AE) and sentiment classification (SC). 
Over the past years, some span-based methods have achieved promising results for ASA, which first extract aspects  by detecting aspect boundaries (AE) and then predict the span-level sentiments (SC).
AE has been widely studied by traditional machine~\cite{jakob2010extracting} and deep learning algorithms~\cite{karimi2021adversarial}. However, the absence of sentiment information may result in redundant and noisy detection.
SC is to predict the sentiment expressed on some specific aspects in a sentence, which has been studied extensively in NLP community~\cite{tang2015effective,chen2019target,chen2020coarse,karimi2021adversarial}.
However, these aspects must be annotated before the AE task. 
 \cite{hu2019open} proposed an extract-then-classify framework, which extracted aspects with a heuristic decoding algorithm and then correspondingly classified the span-level sentiments. \cite{lin2020shared} designed share-private representation for each task to capture the correlations between two tasks in encoding layer. Besides, ~\cite{zhou2019span} introduced a joint model based on span-aware attention mechanism to predict the sentiment polarity.
 Although achieved improved performances,  these works fail to model two-way interactions between AE and SC appropriately.
\section{Conclusion}
In this paper, we proposed a novel aspect-sentiment analysis model named HI-ASA. The proposed model is equipped with a hierarchical interactive network to facilitate information sharing between the aspect extraction task and the sentiment classification task.
The experimental results on three benchmark datasets demonstrated HI-ASA's effectiveness and generality. 
\section*{Acknowledgments}
We are grateful to all the anonymous reviewers for their helpful advice on various aspects of this work. The research work is supported by the National Key Research and Development Program of China under Grant No. 2021ZD0113604, the National Natural Science Foundation of China under Grant No. 62176014, Graduate Research and Innovation Foundation of Chongqing, China under Grant No.CYC21072.

\bibliography{anthology}

\appendix

\end{document}